\documentclass[10pt,twocolumn,letterpaper]{article}

\usepackage{cvpr}              
\usepackage[accsupp]{axessibility}
\usepackage{algorithm} 
\usepackage{algpseudocode} 
\usepackage{ amssymb }
\usepackage[dvipsnames]{xcolor}
\usepackage[dvipsnames]{xcolor}

\definecolor{cvprblue}{rgb}{0.21,0.49,0.74}
\usepackage[pagebackref,breaklinks,colorlinks,citecolor=cvprblue]{hyperref}

\def\paperID{5866} 
\def\confName{CVPR}
\def\confYear{2024}

\title{D3T: Distinctive Dual-Domain Teacher Zigzagging Across RGB-Thermal Gap for Domain-Adaptive Object Detection}

\author{Dinh Phat Do$^{1}$, Taehoon Kim$^{1}$, Jaemin Na$^{1,2}$, Jiwon Kim$^{3}$, Keonho Lee$^{3}$, Kyunghwan Cho$^{3}$,\\ and Wonjun Hwang$^{1}$\\
$^{1}$Ajou University, Korea, $^{2}$Tech. Innovation Group, KT, $^{3}$Robotics Lab, Hyundai Motor Company\\
{\tt\small \{phatai,th951113,wjhwang\}@ajou.ac.kr jaemin.na@kt.com}\\
{\tt\small \{jiwon1115,keonho.lee,kyunghwan.cho\}@hyundai.com}
}

\begin{document}
\maketitle

\begin{abstract}
Domain adaptation for object detection typically entails transferring knowledge from one visible domain to another visible domain. However, there are limited studies on adapting from the visible to the thermal domain, because the domain gap between the visible and thermal domains is much larger than expected, and traditional domain adaptation can not successfully facilitate learning in this situation. To overcome this challenge, we propose a Distinctive Dual-Domain Teacher (D3T) framework that employs distinct training paradigms for each domain. Specifically, we segregate the source and target training sets for building dual-teachers and successively deploy exponential moving average to the student model to individual teachers of each domain. The framework further incorporates a zigzag learning method between dual teachers, facilitating a gradual transition from the visible to thermal domains during training. We validate the superiority of our method through newly designed experimental protocols with well-known thermal datasets, i.e., FLIR and KAIST. Source code is available at \href{https://github.com/EdwardDo69/D3T}{https://github.com/EdwardDo69/D3T}. 
\end{abstract}
\section{Introduction}
\label{sec:intro}

\begin{figure}[t]
  \centering
   \includegraphics[width=1.0\linewidth]{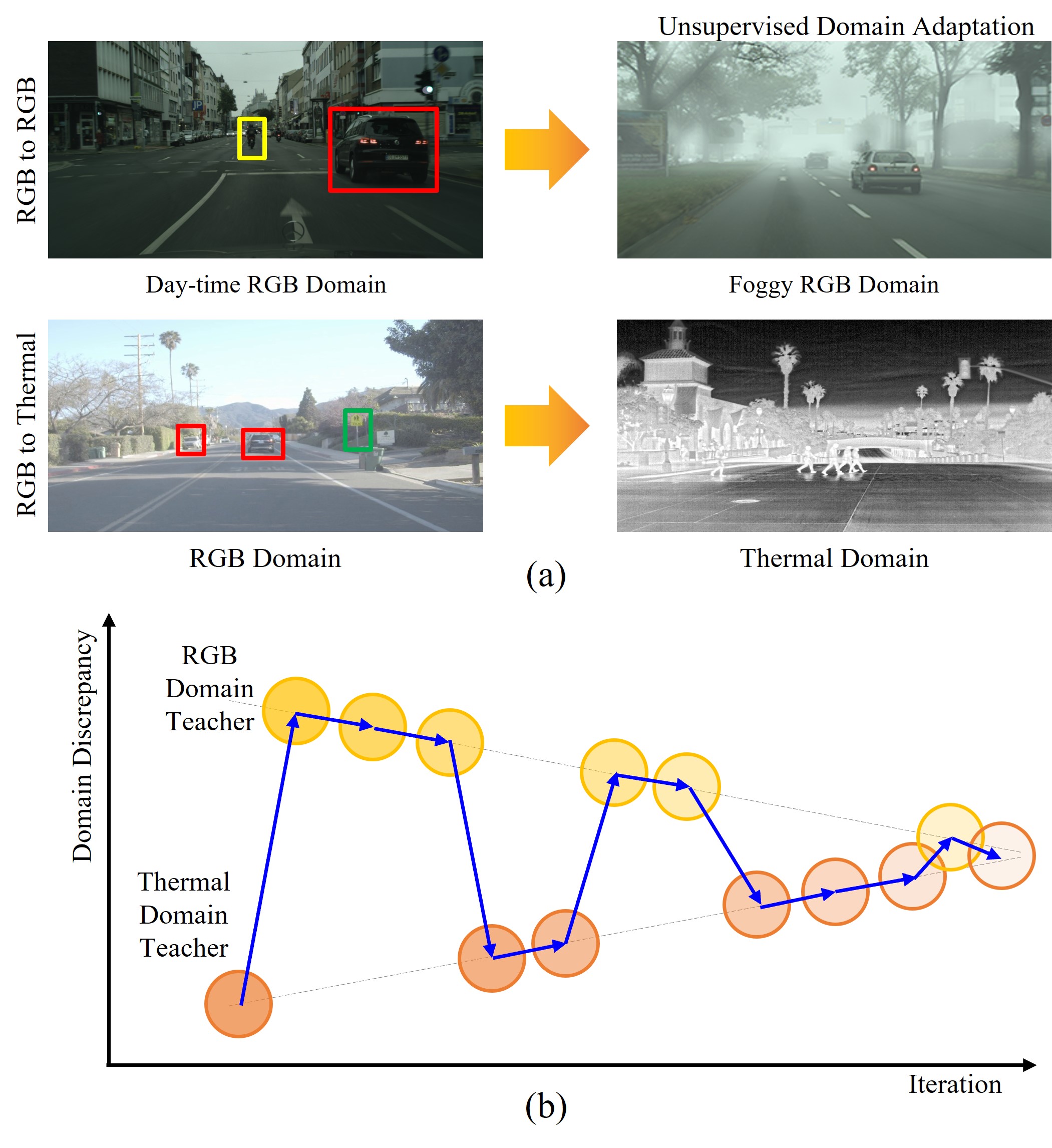}
   \caption{(a) Sample images showing the difference between unsupervised domain adaptation from RGB to RGB domains and unsupervised domain adaptation from RGB to thermal domains. (b) Conceptual illustration of the proposed unsupervised domain adaptation using distinctive dual-domain teachers, demonstrating the zigzag approach across the large RGB-thermal gap.}
   \label{fig:onecol}
   \vspace{-3mm}
\end{figure}

Beyond the significant success of the Convolutional Neural Network (CNN)~\cite{krizhevsky2012imagenet, ResNet}, it has naturally led to recent advancements in CNN-based object detection~\cite{FasterRCNN, yolo, ssd, tian2019fcos}. These advances hold promise for wide real-world applications such as autonomous driving, surveillance, and human activity recognition. Reflecting on the key contributors to this success, two crucial factors emerge: the development of efficient network architectures~\cite{FasterRCNN, yolo} and the availability of a sufficient number of trainable RGB images~\cite{deng2009imagenet, lin2014microsoft, everingham2010pascal} with corresponding supervision signals for supervised learning. It is noteworthy that RGB cameras struggle to provide reliable imaging in scenarios where visible light sensors prove inadequate, particularly during nighttime. In sharp contrast, thermal cameras~\cite{gade2014thermal} hold a significant advantage, detecting the heat emitted by objects and facilitating effective operation in complete darkness, through smoke, and in visually obstructive environments. This capability makes them indispensable for various applications, including nighttime surveillance, search and rescue operations, wildlife monitoring, and all-weather autonomous driving systems~\cite{patel2020night, akshatha2022human, uavod, ippalapally2020object, RealTimeInfrared}.

As we delve into thermal image-based object detection~\cite{akshatha2022human, RealTimeInfrared}, a distinct set of challenges emerges. Foremost among them is the scarcity of annotated thermal datasets essential for training sophisticated detection models. Contrary to the wealth of annotations accessible for RGB object images~\cite{lin2014microsoft}, thermal datasets are notably limited, posing a challenge to the advancement of high-performance thermal detection models using the sufficient training images. The visual features in thermal images diverge significantly from those in RGB images, giving rise to a domain shift problem. This discrepancy leads to performance degradation when models trained on visible datasets are applied to thermal images. Consequently, addressing these challenges necessitates the employment of specialized training and adaptation techniques to construct effective object detection systems capable of harnessing the unique properties inherent in thermal cameras. 

In this paper, we leverage Unsupervised Domain Adaptation (UDA) methods~\cite{ganin2015unsupervised,tzeng2017adversarial,xie2018learning} to alleviate the domain shift problems from the source domain (e.g., RGB images) to the target domain (e.g., thermal images). We have focused on UDA for object detection~\cite{chen2018domain, saito2019strong}. This aims to minimize the discrepancy between source and target domains and enhance model performance without requiring labor-intensive labeling of target data. We have focused to \emph{one-stage object detection} method, e.g., FCOS~\cite{tian2019fcos} in this paper, because it is generally faster than two-stage object detection for real-time applications. This is particularly crucial in applications such as autonomous driving, where acquiring labeled thermal images can be both time-consuming and expensive. While the aforementioned methods primarily utilize conventional UDA methods based on only RGB images, they fall short in addressing the fundamental challenge of UDA from RGB to thermal images. As shown in Fig.~\ref{fig:onecol} (a), it stems from the significant disparity between the RGB and thermal domains compared to that between two RGB domains. 

To solve this issue, we propose a novel Mean Teacher (MT) framework using Distinctive Dual-Domain Teacher (D3T) for domain adaptive object detection between RGB and thermal domains. Unlike prior MT-based object detections (e.g., single teacher and single student)~\cite{deng2021unbiased, li2022cross}, we employ two distinct teacher models, each specializing in either RGB or thermal domain. This facilitates more effective learning of domain-specific information, particularly in the presence of the substantial discrepancy. This D3T framework, paired with a zigzag learning method (as shown in Fig.~\ref{fig:onecol} (b)) between domains, updates selected domain-specific weights to the single student, enabling a gradual transition from RGB to thermal domains. By zigzagging the teacher network selection, we leverage the observation that, during initial training, the RGB teacher pre-trained from source labels is more likely to predict relatively accurate pseudo-labels on the target, while the thermal teacher performs better as training progresses. To achieve this, we adjust the selection frequency, favoring the RGB teacher more in the early stages of training and gradually increasing the emphasis on the thermal teacher as training progresses. Finally, we verify performances of our method using new established evaluation protocols with well-known thermal datasets such as FLIR~\cite{zhang2020multispectral} and KAIST~\cite{hwang2015multispectral}. 

We summarize our contributions as follows:
\begin{itemize}
    \item We introduce the D3T framework, leveraging two distinctive domain teachers for effective domain adaptive object detection between RGB and thermal domains.  
\end{itemize}
\begin{itemize}
    \item Our zigzag learning method facilitates a gradual shift from RGB to thermal domains, updating domain-specific weights dynamically. This optimizes adaptation, leveraging each teacher's strengths during training.
\end{itemize}
\begin{itemize}
    \item We have made our experimental protocols using well-known thermal datasets: FLIR and KAIST, and prove the superiority of our method compared with other methods.
\end{itemize}

\section{Related Work}
\label{sec:related}
\subsection{Thermal Object Detection}
Thermal object detection~\cite{akshatha2022human} is pivotal for applications in surveillance, military operations, and autonomous driving. Recent advancements underscore their adaptability and efficiency~\cite{RealTimeInfrared, uavod, ippalapally2020object, yao2022infrared}. A notable trend is the fusion of visible and thermal features, enhancing detection accuracy by capturing more comprehensive environmental information~\cite{chen2022multimodal, zhang2021guided, zhou2020improving, cao2023multi}. 
However, these studies typically assume the simultaneous capture of visible and thermal images and they should be aligned well.

\subsection{UDA for Object Detection}
UDA for object detection is focusing on adapting detectors from a labeled source domain to an unlabeled target domain. The primary methods in UDA are categorized into domain alignment and self-training. Domain alignment techniques including style transfer~\cite{chen2020harmonizing, kim2019diversify, inoue2018cross}, adversarial training~\cite{chen2018domain, saito2019strong, hsu2020every}, and graph matching~\cite{li2022sigma, xu2020cross, zhang2021rpn} aim to minimize the domain discrepancy by aligning features or visual styles between the source and target domains. However, these methods face challenges in maintaining a balance between feature transferability and discriminability. In contrast, self-training methods leverage inherent information from the target domain. The UMT~\cite{deng2021unbiased} generates pseudo labels using similar images to the source domain, while the HT~\cite{deng2023harmonious} emphasizes consistency in classification and localization, using a new sample reweighting scheme. The unified CMT~\cite{cao2023contrastive} framework employs self-training with contrastive learning in domain-adaptive object detection. This enhances target domain performance by optimizing object-level features using pseudo-labels without requiring target domain labels. UDA is crucial for enabling object detection models to perform accurately across diverse environments, especially where labeling data in the target domain is impractical, such as in autonomous driving at night.

\subsection{Domain Adaptive Thermal Object Detection}
Domain adaptive thermal object detection aims to enhance object detection in thermal images, especially in suboptimal lighting conditions. This field addresses the limitations inherent in object detectors designed for visible light datasets, which typically underperform in environments with poor or variable lighting. Utilizing UDA, these methods leverage labeled data from the visible spectrum to improve detection in the thermal spectrum with the limited availability of labeled thermal data. Despite its considerable potential for practical applications, this field currently attracts relatively modest research investment. Meta-UDA approach~\cite{vs2022meta} stands out as a significant advancement by leveraging an algorithm-agnostic meta-learning framework for better domain adaptation using labeled data from visible domains. Nakamura et al.~\cite{nakamura2022few} introduces a unique data fusion strategy using CutMix. This approach integrates elements of target images into source images, coupled with adversarial learning, resulting in enhanced object detection efficacy.

The previous methods only take advantage of UDA for RGB images and it is not easy to bridge the large gap between the RGB and thermal domains. To overcome this, we propose the D3T framework collaborated with a zigzag learning method specifically designed from RGB to thermal domain adaptation, which is highly efficient and easy to implement for domain adaptive object detection.

\section{Proposed Method}
\label{sec:proposed}
In the quest for advancing object detection capabilities across diverse imaging domains, we delve into the Mean Teacher (MT) framework~\cite{tarvainen2017mean} and extend it with a dual teacher-based framework. 

\begin{figure*}[t]
  \centering
   \includegraphics[width=1.0\linewidth]{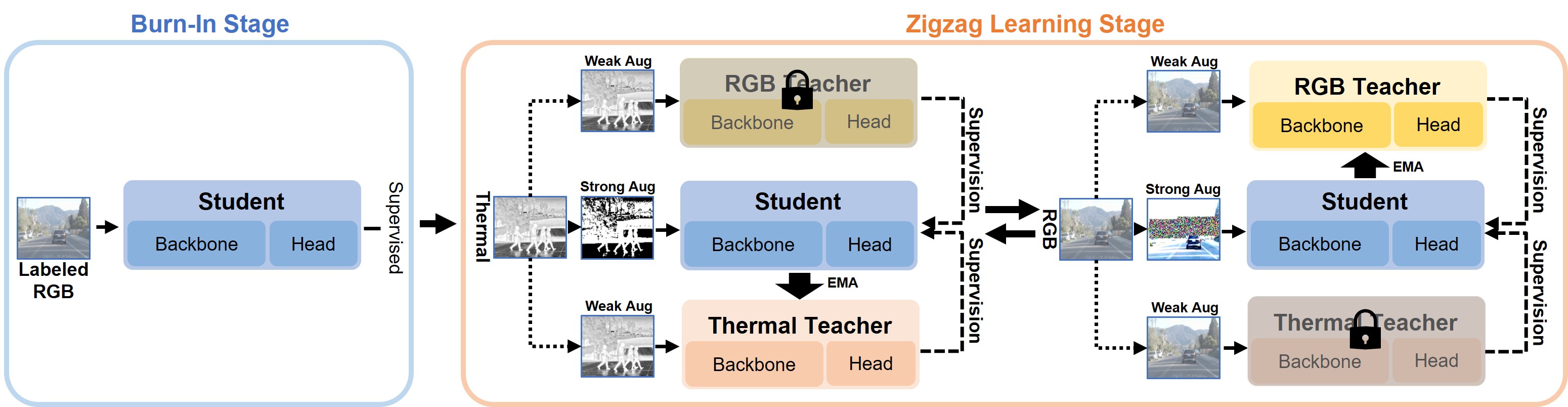}
  \caption{Overview of \textbf{D3T}: Our D3T model consists of two stages. \textbf{Burn-in Stage}: We initiate the training of the object detector using labeled data from the RGB domain. \textbf{Zigzag Learning Stage}: Comprises two distinct and interleaved training components for the Thermal domain and the RGB domain, respectively. During each step of training, the student model utilizes images from a single domain for training but leverages knowledge from two teachers for enhanced learning effectiveness. In each step, only one teacher model is updated corresponding to the trained domain.}
  \label{fig:twocol}
  \vspace{-3mm}
\end{figure*}

\subsection{MT Framework with A Single Teacher}
The MT framework represents a paradigm in domain adaptation, particularly within the context of object detection tasks~\cite{deng2021unbiased, li2022cross, deng2023harmonious}. This approach learns knowledge from labeled data in the source domain and adapts it to the unlabeled target domain. Furthermore, it employs the teacher-student mutual learning method, as introduced in~\cite{liu2021unbiased}, to enhance detection accuracy.

\textbf{Overview:} The core idea of the MT framework is a model architecture consisting of a teacher model and a student model, two detectors with identical architectures. The teacher model, pre-trained on labeled data from the source domain, generates pseudo-labels for the target domain data, which lacks labels. The student model is optimized by using these pseudo-labels, and its weights are updated to the single teacher model. The teacher model can be regarded as the ensemble of student models at various time steps, resulting in higher accuracy and the production of better quality pseudo labels.

\textbf{Training method: }The MT framework uses both source and target domains for training at the same time. The source domain data is applied with both strong and weak data augmentation before being directly used for the supervised training of the student model with ground-truth labels. Target domain data employs two types of data augmentation: weak augmentation for the teacher model's input images to ensure reliable pseudo-labels, and strong augmentation for the student model's input images to enhance the model's diversity. This enhances the teacher model since it is updated with the weights from the student model at various time steps.

The overall loss function for the MT framework is defined as follows:
\begin{equation}
    \mathcal{L} = \mathcal{L}_{src} + \mathcal{L}_{tgt}, 
    \label{eq:01}
\end{equation}
where $\mathcal{L}_{src}$ is the loss in the source domain, including classification and localization loss, and $\mathcal{L}_{tgt}$ is the loss in the target domain which is similarly calculated using pseudo-labels.

\textbf{Update teacher parameter:} The MT framework updates the weights of the teacher model with the weights of the student model via Exponential Moving Average (EMA). This gradual updating process results in the teacher model becoming an ensemble of student models across different time steps and it is derived by
\begin{equation}
    \theta^\mathcal{T} \leftarrow \alpha\theta^\mathcal{T} + (1 - \alpha)\theta^\mathcal{S}.
    \label{eq:02}
\end{equation}
where $\theta^\mathcal{T}$ represents the weights of the teacher model, $\theta^\mathcal{S}$ represents the weights of the student model, and $\alpha$ is the EMA coefficient. For simplicity, we set $\alpha$ as 0.9996 in all experiments.

\subsection{Distinctive Dual-Domain Teacher (D3T)} \label{sec:3.2}
UDA for object detection typically employs an MT framework with a single teacher model to adapt across RGB image domains, such as from the Cityscapes~\cite{cordts2016cityscapes} to the Foggy Cityscapes dataset~\cite{sakaridis2018semantic}. However, the domain gap between the RGB and thermal domains is significantly larger. Therefore, using a single teacher model for both domains can lead to negative effects and diminish the model's effectiveness. To address this issue, we introduce a new initiative called D3T, which is directly inspired by~\cite{Na2023} and includes two individual-teacher models for the RGB domain and the thermal domain, respectively. The two teacher models leverage the specialized knowledge of their respective domains and transfer this knowledge to the student model. The overview of D3T is summarized in Fig.~\ref{fig:twocol}.

\textbf{Separate teachers:} The core idea of our method is to use two separate teachers, an RGB teacher and a thermal teacher, to integrate knowledge from their respective domains. Each teacher's model is updated with the student model's weights only when it is trained with the corresponding domain. As a result, that teacher acquires the specialized knowledge of that domain without being negatively impacted by other domains. The D3T model is trained using thermal images and updates the weights for the corresponding thermal teacher. Similar to the right side, our model is trained with the RGB domain and updates the weights for the RGB teacher model.

\textbf{Learning knowledge from Dual-Teachers:} During each training step of the D3T model, images from only one domain, either RGB or thermal domain, are used. However, to leverage the combined knowledge of both teachers and minimize the domain shift between the two domains, both thermal and RGB teachers are employed to generate pseudo-labels. The dual teaching method not only utilizes the knowledge from the two teachers but also increases the reliability of the pseudo-labels, which leads to more effective training of the student model. The loss functions are defined as follows:
\begin{equation} 
    \mathcal{L}_{rgb\_sup} = \mathcal{L}_{sup}(f^\mathcal{S}(\mathcal{I}_{rgb}), \mathcal{Y}),
    \label{eq:03}
\end{equation}
\begin{equation} 
\begin{split}
    &\mathcal{L}_{thr} =\\ &\mathcal{L}_{un}(f^\mathcal{S}(\mathcal{I}_{thr}), f^\mathcal{T}_{thr}(\mathcal{I}_{thr}))
    + \mathcal{L}_{un}(f^\mathcal{S}(\mathcal{I}_{thr}), f^\mathcal{T}_{rgb}(\mathcal{I}_{thr})),
\end{split}
\label{eq:04}
\end{equation}
where, $\mathcal{L}_{thr}$ is the loss for the thermal domain, and $\mathcal{L}_{rgb\_sup}$ represents the supervised loss for the RGB domain. Similarly, $\mathcal{I}_{thr}$ and $\mathcal{I}_{rgb}$ denote the images from thermal and RGB domains, respectively. $f^S$ corresponds to the student model, which generates predictions for the input images. Whereas $f^\mathcal{T}_{thr}$ and $f^\mathcal{T}_{rgb}$, representing the teacher models for the thermal and RGB domains, are responsible for generating pseudo-labels to train the student model. $\mathcal{Y}$ denotes the ground truth labels for the images in the RGB source domain. The losses consist of the unsupervised loss $\mathcal{L}_{un}$ and the supervised loss $\mathcal{L}_{sup}$, which are used like~\cite{deng2023harmonious}.

\subsection{Zigzag Learning Across RGB-Thermal Domains} \label{sec:3.3}
In traditional UDA methods for object detection, the source and target domains were commonly trained simultaneously. However, due to the substantial domain gap between the RGB and thermal domains, simultaneously training is ineffective. We propose a training approach for domain adaptation from RGB to thermal, which is called zigzag learning. 

\textbf{Distinctive training:} The zigzag learning involves separate and alternate training for the RGB and thermal domains to learn the distinct knowledge of each domain effectively. Each time we train a specific domain, we update the weights to the teacher model of the corresponding domain using EMA. This domain specific training and weight updating strategy ensures that the significant domain gap between the RGB and thermal domains does not result in negative cross domain influence.

\textbf{Progressive training transition:} The concept of the zigzag learning method is a progressive training transfer process that starts with a focus on learning knowledge from the labeled RGB domain. Next, the training progressively transitions to the thermal domain by steadily increasing the training frequency for thermal images and simultaneously reducing the training frequency for RGB images. This gradual shift facilitates a smooth domain adaptation from the RGB to the thermal domain, resulting in improved performance within the thermal domain. As illustrated in Fig.~\ref{fig:onecol} (b), for example, the unlabeled thermal domain is trained a single time at first, while the labeled RGB domain is trained three times to focus on acquiring knowledge from the RGB domain. Subsequently, the frequency of training sessions in the RGB domain is decreased, while it is increased for the thermal domain, facilitating domain adaptation from RGB to thermal between the two domains. The training iterations for the RGB and thermal domains at each step are defined as follows:

\begin{equation} 
\begin{split}
    Z^{t}_{thr} = Z^{t-1}_{thr} + \beta, \\
    Z^{t}_{rgb} = Z^{t-1}_{rgb} - \beta,
    \label{eq:08}
\end{split}
\end{equation}
where $Z^{t}_{\text{thr}}$ and $Z^{t}_{\text{rgb}}$ are the number of training iterations for the thermal domain and RGB domain at $t^{th}$ step, $Z^{t-1}_{\text{thr}}$ and $Z^{t-1}_{\text{rgb}}$ are the number of training iterations at $({t-1})^{th}$ step. $\beta$ indicates the number of iterations that are adjusted after each step. This equation guarantees that the sum of $Z^{t}_{\text{thr}}$ and $Z^{t}_{\text{rgb}}$ remains constant, while the ratio $Z^{t}_{\text{thr}}:Z^{t}_{\text{rgb}}$ increases incrementally at each step. We have presented a pseudocode of the zigzag learning algorithm in Algorithm~\ref{alg:01}.

\begin{algorithm}[t]
    \caption{Zigzag learning method} 
    \begin{algorithmic}
    
    \Require  
    \State $I$: Total number of iterations,
    $\alpha$: EMA coefficient, $Z_{thr}$, $Z_{rgb}$: Training iterations of thermal and RGB at each step, 
 $\theta^\mathcal{S}$,$\theta^\mathcal{T}_{thr}$,$\theta^\mathcal{T}_{rgb}$: Weights of student, thermal and RGB teachers, $\mathcal{I}_{thr}$, $\mathcal{I}_{rgb}$:  Input of thermal and RGB images.
    \Ensure
    \State $switch \leftarrow Z_{thr}$
    \For{iteration $i \in \{0, 1, 2, \ldots, I$\}}
        \If {$i < switch$} \Comment{Update only thermal teacher}
        \State Calculate $\mathcal{L}_{thr}$ by  $\mathcal{I}_{thr}$
        \State $\theta^\mathcal{S} \leftarrow \mathcal{L}_{thr}$ 
        \State $\theta^\mathcal{T}_{thr} \leftarrow \alpha\theta^\mathcal{T}_{thr} + (1 - \alpha)\theta^\mathcal{S}$ 
        \Else \Comment{Update only RGB teacher}
        \State Calculate $\mathcal{L}_{rgb}$ by  $\mathcal{I}_{rgb}$
        \State $\theta^\mathcal{S} \leftarrow \mathcal{L}_{rgb}$ 
        \State $\theta^\mathcal{T}_{rgb} \leftarrow \alpha\theta^\mathcal{T}_{rgb} + (1 - \alpha)\theta^\mathcal{S}$ 
        \EndIf
        \If {$i >0$ \text{and} $i \% (Z_{thr}+Z_{rgb})==0$ } 
        \State $switch \leftarrow switch + Z_{thr} + Z_{rgb}$
        \EndIf
    \EndFor
    \end{algorithmic} 
    \label{alg:01}
\end{algorithm}

\subsection{Incorporating Knowledge from Teacher Models} \label{sec:3.4}
Our experiments on the effectiveness of domain adaptation techniques indicate some limitations when training within the RGB domain using only ground truth labels. In this section, we describe the limitations and propose an improved strategy that integrates pseudo-labels to enhance knowledge transfer.

\textbf{Limitations of training with only ground truth labels:} We found that training the student model using only ground truth labels poses challenges because the complexity of the ground truth labels makes it difficult for the student model to learn effectively from strongly augmented input images. This leads us to our first observation: a combination of ground truth and pseudo-labels is more effective for knowledge transfer from the teacher model to the student model than training with only ground truth labels. This combination makes the process of transferring knowledge from the teacher model to the student model more effective.

Secondly, we find that training solely with ground truth labels from the RGB domain does not utilize the knowledge synthesized by the thermal teacher model, thereby reducing the effectiveness of domain adaptation from the RGB to the thermal domain. To address these issues, we strategically integrate pseudo-labels generated by both the RGB and thermal teacher models, as well as ground truth labels, into the training for the RGB domain.

\textbf{Pseudo label integration:} However, the direct use of pseudo-labels leads to poor results. The experiments detailed in Table~\ref{tab:06} indicate that using pseudo-labels in the same manner as ground truth labels (with \(\lambda\) equals $1$ ) results in a substantial decline in model performance. As in Section~\ref{sec:3.3}, our method initially focuses on training with ground truth labels, and then we gradually integrate the pseudo-labels from both teachers, alongside the ground truth labels, into the training process. This approach is defined by the following set of equations:
\begin{equation} 
\begin{split}
&\mathcal{L}_{rgb\_unsup} =\\ &\mathcal{L}_{un}(f^\mathcal{S}(\mathcal{I}_{rgb}), f^\mathcal{T}_{rgb}(\mathcal{I}_{rgb})) 
+ \mathcal{L}_{un}(f^\mathcal{S}(\mathcal{I}_{rgb}), f^\mathcal{T}_{thr}(\mathcal{I}_{rgb})),
\label{eq:5}
\end{split}
\end{equation}
\begin{equation} 
\mathcal{L}_{rgb} = \mathcal{L}_{rgb\_sup} + \lambda \mathcal{L}_{rgb\_unsup}.
\label{eq:6}
\end{equation}

In this equation, \(\lambda\) is a hyperparameter that controls the degree to which pseudo-labels are used during training in the RGB domain. This hyperparameter is employed to balance the influence of pseudo-labels and ensure that the student model benefits from the knowledge provided by the teacher models without negative effects. The unsupervised loss \(\mathcal{L}_{un}\) is utilized in a similar manner as in Section~\ref{sec:3.2}.

The total loss for the D3T model is formulated as follows:
\begin{equation} 
\mathcal{L}_{all}=
\begin{cases}
\mathcal{L}_{thr} & \text{training with thermal domain},\\
\mathcal{L}_{rgb} & \text{training with RGB domain}.
\end{cases}
\end{equation}

\section{Experimental Results and Discussions}
\label{sec:exp}

\subsection{Dataset and Evaluation Protocol}
We evaluate our proposed method using the following datasets and new designed domain adaptation evaluation protocols from RGB to thermal domains;

\textbf{FLIR~\cite{zhang2020multispectral}:} In our research, we chose the updated FLIR dataset over the older one~\cite{fa2018flir} because it has many labeling errors. The dataset includes 5,142 precisely aligned pairs of color and infrared images, with 4,129 used to train our method and 1,013 used for testing it. These images are from the view of a car driver and include both daytime and nighttime scenes. We are only looking at objects like ``people," ``cars," and ``bicycle" that have complete labels to make sure our evaluation is accurate.

\textbf{KAIST~\cite{hwang2015multispectral}:} The renowned KAIST dataset comprises 95,328 pairs of color and thermal images. We employ an updated version with more precise labeling as provided by~\cite{zhang2019weakly}. This version contains 8,892 accurately adjusted pairs of RGB-Thermal images for training and 2,252 pairs for evaluation purposes.

\textbf{RGB$\rightarrow$Thermal FLIR evaluation:} The FLIR dataset, known for its precisely aligned image pairs, can cause models to overfit and may not accurately reflect the true performance of domain adaptation algorithms. To address this, we introduce a disjointed image training approach. We use the first 2,064 RGB images as the source domain and a separate set of 2,064 thermal images as the target domain for training. Note that RGB source and thermal target images are exclusively selected. This method guarantees that the training does not use any matching RGB-Thermal image pairs, preventing overfitting and providing a more reliable assessment of the domain adaptation algorithm's effectiveness.

\textbf{RGB$\rightarrow$Thermal KAIST evaluation:} Like with the FLIR dataset, we apply a disjointed image training approach for the KAIST dataset. We select the initial 4,446 RGB images as the source domain and the subsequent 4,446 thermal images as the target domain, ensuring that training does not involve any matched image pairs. RGB source and thermal target images are exclusively selected. Furthermore, we have removed any images without labels, resulting in a total of 1,216 images to validate the algorithm's performances. 

\subsection{Implemental Details}
Following the baseline~\cite{deng2023harmonious}, we deploy the FCOS detector~\cite{tian2019fcos} equipped with a VGG-16 backbone for the FLIR dataset and a ResNet-50 backbone for the KAIST dataset in our experiments. Our experiments run on a batch size of 8 using 4 NVIDIA RTX A5000 GPUs. In accordance with~\cite{deng2023harmonious}, we initiate the learning rate at 0.005 and not apply any decay. For data augmentation, we adopt the same strategy as in~\cite{deng2023harmonious, liu2021unbiased}, resizing the shortest edge of images to a maximum of 800 pixels.
In Section~\ref{sec:3.3}, for the FLIR dataset, \( Z^0_{\text{thr}} \) and \( Z^0_{\text{rgb}} \) are initialized to 50 and 150, respectively. They will be adjusted every 10k iterations by a \( \beta \) value of 50 as specified in the equation ~(\ref{eq:08}). For the KAIST dataset, \( Z^0_{\text{thr}} \), \( Z^0_{\text{rgb}} \), and \( \beta \) are initially established at 25, 75, and 25 respectively, and each adjustment step comprises 10k iterations.

\subsection{Performance Comparison Table}
We compare our proposed method with the well-known domain adaptation methods. 

\textbf{RGB$\rightarrow$Thermal FLIR evaluation:} The adaptation results for RGB to thermal image conversion on the FLIR dataset, as presented in Table~\ref{tab:01}, indicate that our D3T method has achieved remarkable performance, surpassing other advanced technologies in domain adaptation. Specifically, the D3T method outperforms the HT~\cite{deng2023harmonious} algorithm, which is a significant player in the field utilizing a student-teacher framework, by 3.49\% in mean Average Precision (mAP). Notably, HT~\cite{deng2023harmonious} itself had previously set a high benchmark by outperforming the EPM~\cite{hsu2020every} method, which does not use the student-teacher approach, by 21.21\% in mAP.

The advancements observed in our study highlight a critical insight: previous algorithms have not adequately tackled the considerable domain gap between RGB and thermal domains. This gap poses a more formidable challenge than those encountered in typical adaptation scenarios, such as transitioning from Cityscapes to Foggy Cityscapes. Our experimental results unequivocally showcase the efficacy of the proposed D3T method in effectively addressing this issue, marking a substantial leap forward in domain adaptation.

\begin{table}[t]
  \centering
  \small
      \begin{tabular}{c|ccc|c}
        \hline
        Method & Person & Bicycles & Car & mAP\\
        \hline
        \hline
        Source only & 28.54 & 28.28 & 47.22 & 34.68\\
        DANN~\cite{ganin2016domain} & 32.02 & 30.52 & 48.88 & 37.14\\
        SWDA~\cite{saito2019strong} & 30.91 & 36.03 & 47.94 & 38.29\\
        EPM~\cite{hsu2020every} & 40.97 & 38.95 & 53.83 & 44.60\\
        HT~\cite{deng2023harmonious} & \textbf{70.87} & \underline{48.11} & \underline{78.45}& \underline{65.81}\\
        \hline
        D3T(Ours) & \underline{70.77} & \textbf{57.44} & \textbf{79.68} & \textbf{69.30}\\
        \hline
      \end{tabular}
  \vspace{-3mm}
  \caption{Adaptation results of FLIR dataset from RGB images to thermal images with VGG16 backbone. The best accuracy is indicated in bold, and the second-best accuracy is underlined.}
  \label{tab:01}
  \vspace{-3mm}
\end{table}

\textbf{RGB$\rightarrow$Thermal KAIST evaluation:} The domain adaptation results for converting RGB images to thermal images on the KAIST dataset, as depicted in Table~\ref{tab:02}, demonstrate the superior performance of our D3T algorithm. D3T outperforms the HT~\cite{deng2023harmonious} algorithm, which is one of the most advanced algorithms in this domain, by a significant margin of 5.51\% in mAP. Furthermore, when compared to the EPM~\cite{hsu2020every} algorithm, which does not utilize a student-teacher framework, the D3T method shows an even greater improvement of 9.41\% mAP. This impressive advancement illustrates the effectiveness of the D3T algorithm in addressing the challenges of domain adaptation from RGB to thermal domain.

\begin{table}[t]
  \centering  
  \small
  \begin{tabular}{c|c}
    \hline
    Method & Person\\
    \hline
    \hline
        RGB Source only & 9.09\\
        DANN~\cite{ganin2016domain} & 9.17\\
        SWDA~\cite{saito2019strong} & 31.30\\
        EPM~\cite{hsu2020every} & 39.55\\
        HT~\cite{deng2023harmonious} & \underline{43.45}\\
        \hline
        D3T(Ours) & \textbf{48.96}\\
        \hline
  \end{tabular}
  \vspace{-3mm}
  \caption{Adaptation results on KAIST dataset from RGB images to thermal images with Resnet-50 backbone. The best accuracy is indicated in bold, and the second-best accuracy is underlined.}
  \label{tab:02}
  \vspace{-3mm}
\end{table}

\subsection{Ablation Experiments}
We make ablations and detail discussions in this section.

\begin{figure}[t]
  \centering
   \includegraphics[width=1.0\linewidth]{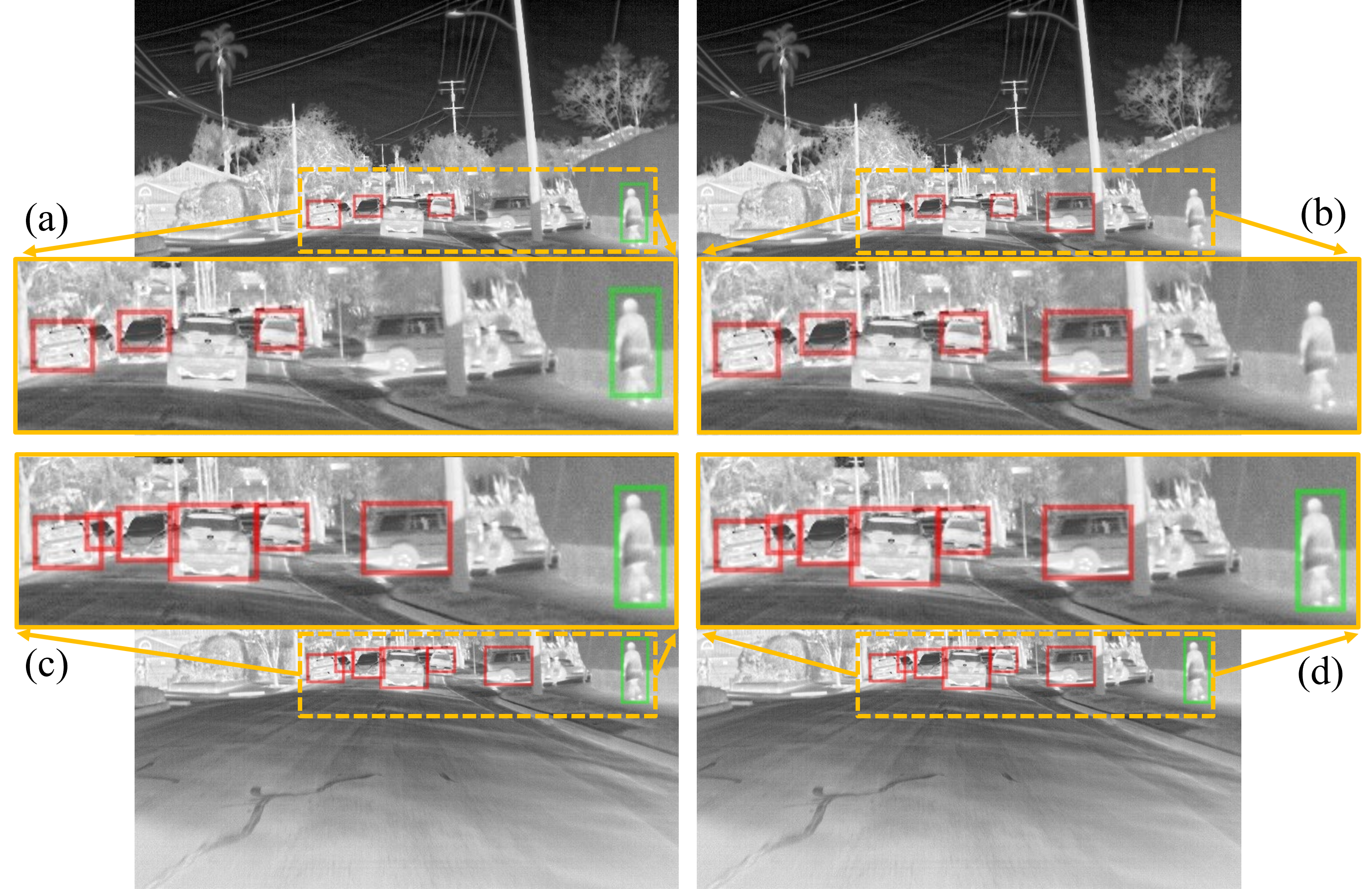}
   \vspace{-3mm}
   \caption{Dual-teachers' pseudo-labels at different training stages. (a) and (b) are pseudo-labels from the RGB and thermal teacher models in early training stages, respectively, while (c) and (d) are pseudo-labels from the same models in later training stages.}
   \label{fig:05}
   \vspace{-3mm}
\end{figure}

\begin{figure*}
     \centering
     \begin{subfigure}[b]{0.245\textwidth}
         \centering
         \includegraphics[width=\textwidth]{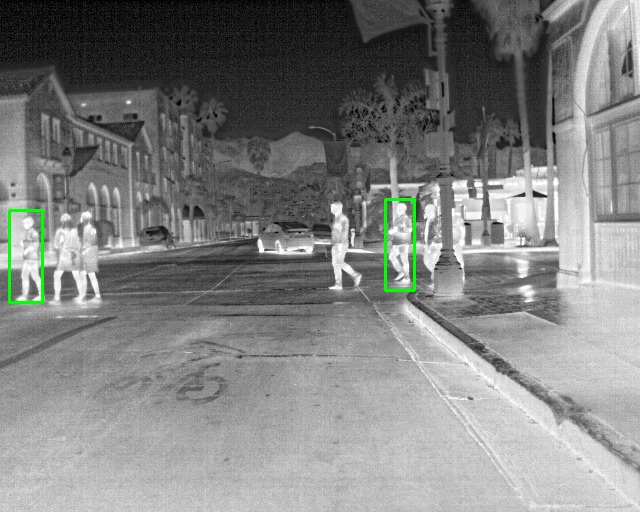}
         \caption{Source only}
     \end{subfigure}
     \hfill
     \begin{subfigure}[b]{0.245\textwidth}
         \centering
         \includegraphics[width=\textwidth]{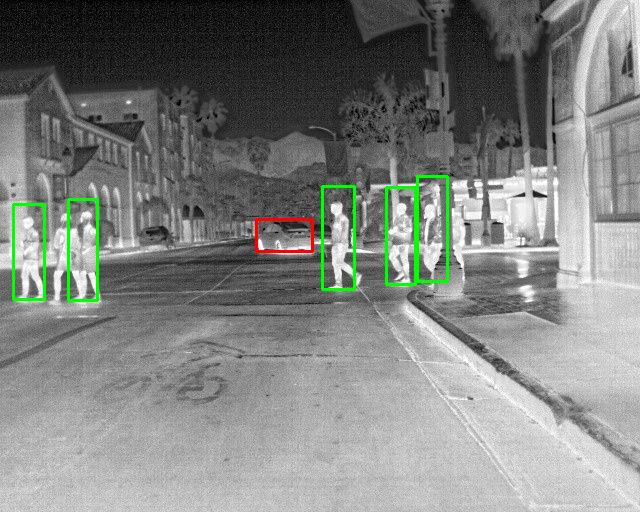}
         \caption{EPM~\cite{hsu2020every}}
     \end{subfigure}
     \hfill
     \begin{subfigure}[b]{0.245\textwidth}
         \centering
         \includegraphics[width=\textwidth]{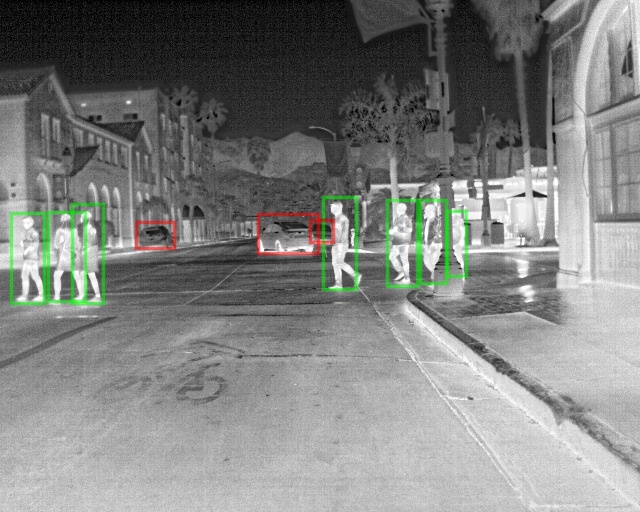}
         \caption{D3T (Ours)}
     \end{subfigure}
     \hfill
     \begin{subfigure}[b]{0.245\textwidth}
         \centering
         \includegraphics[width=\textwidth]{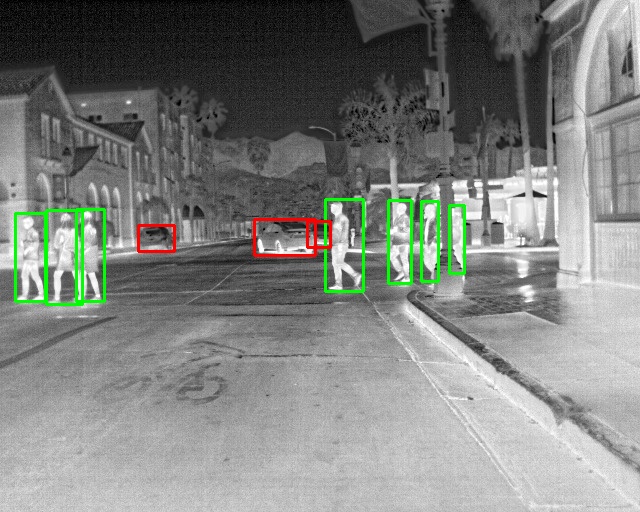}
         \caption{Ground truth}
     \end{subfigure}
        \vspace{-3mm}
        \caption{Visualization of UDA results for object detection models: Source only, EPM~\cite{hsu2020every}, our D3T, and ground truth labels in the FLIR dataset RGB $\rightarrow$ thermal domain. The \textcolor{green}{green} and \textcolor{red}{red} boxes represent the classes of \textcolor{green}{person} and \textcolor{red}{car}.}
        \label{fig:03}
        \vspace{-3mm}
\end{figure*}

\begin{figure*}
     \centering
     \begin{subfigure}[b]{0.245\textwidth}
         \centering
         \includegraphics[width=\textwidth]{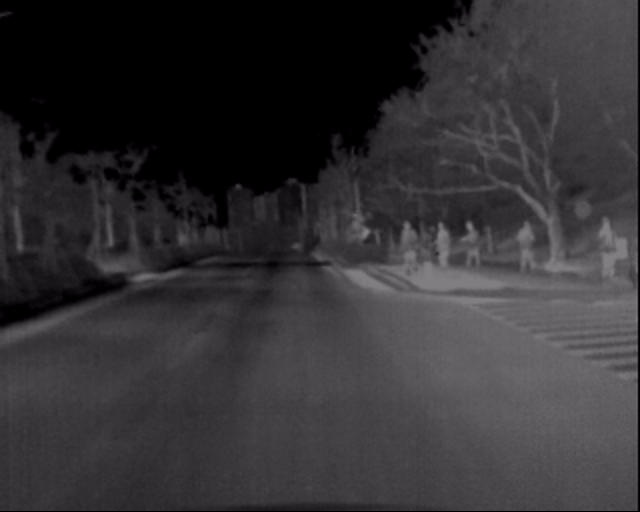}
         \caption{Source only}
     \end{subfigure}
     \hfill
     \begin{subfigure}[b]{0.245\textwidth}
         \centering
         \includegraphics[width=\textwidth]{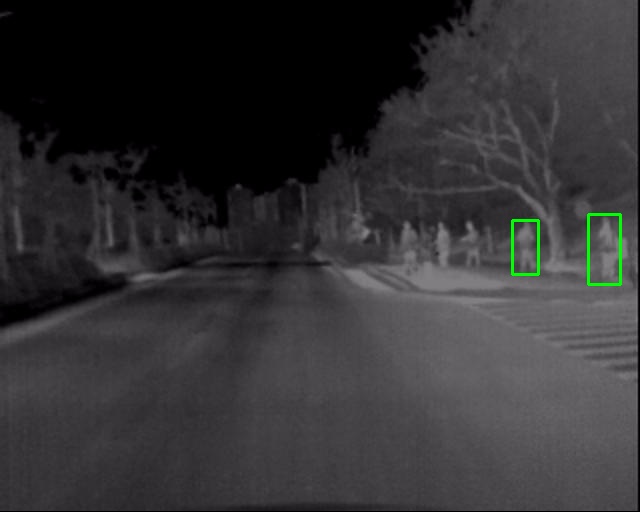}
         \caption{EPM~\cite{hsu2020every}}
     \end{subfigure}
     \hfill
     \begin{subfigure}[b]{0.245\textwidth}
         \centering
         \includegraphics[width=\textwidth]{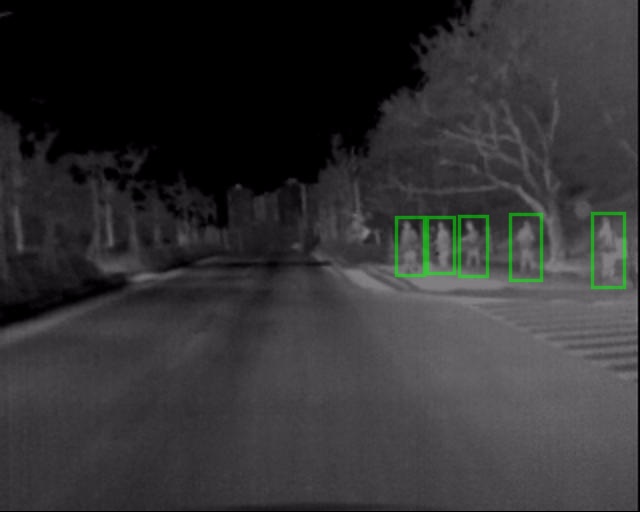}
         \caption{D3T (Ours)}
     \end{subfigure}
     \hfill
     \begin{subfigure}[b]{0.245\textwidth}
         \centering
         \includegraphics[width=\textwidth]{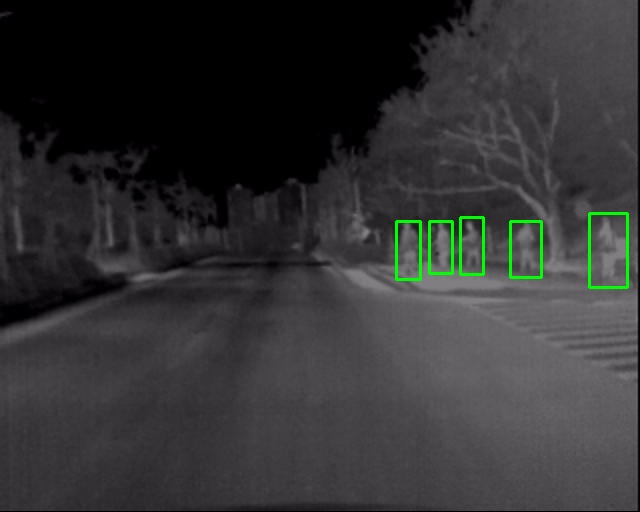}
         \caption{Ground truth}
     \end{subfigure}
     \vspace{-3mm}
        \caption{Visualization of UDA results for object detection models: Source only, EPM~\cite{hsu2020every}, our D3T, and ground truth labels in the KAIST dataset RGB $\rightarrow$ thermal domain. The \textcolor{green}{green} boxes represent the classes of \textcolor{green}{person}.}
        \label{fig:04}        
        \vspace{-3mm}
\end{figure*}

\textbf{Visualization:} Fig.~\ref{fig:05} illustrates the effectiveness of our D3T model. At the early training steps, each teacher holds specific knowledge relevant to their respective domain. Therefore, the teachers created different pseudo labels as illustrated in Fig.~\ref{fig:05} (a) and (b). In the final training steps, the two teachers provided pseudo labels of high quality that were similar. This indicates that our D3T algorithm improves model efficiency and bridges the domain gap. This corresponds to our concept presented in Fig.~\ref{fig:onecol} (b). We also show object detection results on two datasets, FLIR and KAIST, in Fig.~\ref{fig:03} and Fig.~\ref{fig:04} to provide a visual comparison of the effectiveness of our D3T method.

\textbf{Ablation study:} The ablation study for the D3T model, as detailed in Table~\ref{tab:03}, assesses the adaptation from RGB to thermal images on the FLIR dataset and sheds light on the contributions of various model components. Starting from a baseline mAP of 65.81\%, the addition of Dual-Teachers provides a notable improvement, bringing the mAP to 66.93\%. Further incorporation of Zigzag-Learn with Dual-Teachers enhances the mAP marginally to 68.46\%, suggesting the effectiveness of alternating training strategies in domain adaptation. The full model, integrating Dual-Teachers, Zigzag-Learn, and Incor-Know, achieves the most significant performance leap, culminating in an mAP of 69.30\%. This comprehensive approach highlights the synergistic impact of combining domain-specific knowledge acquisition, specialized training methods, and robust cross-domain knowledge integration to effectively adapt RGB image detection models for thermal image applications.

\begin{table}[t]
  \centering
  \small
  \begin{tabular}{ccc|c}
    \hline
    Dual-Teachers & Zigzag-Learn & Incor-Know & mAP\\
    \hline
    \hline
     &  &  & 65.81\\
    \checkmark &  &  & 66.93\\
    \checkmark & \checkmark &  & 68.46\\
    \checkmark & \checkmark &\checkmark & \textbf{69.30}\\
    \hline
  \end{tabular}
  \vspace{-3mm}
  \caption{Ablation studies of D3T on FLIR dataset from RGB images to thermal images. Dual-Teachers, Zigzag-Learn and Incor-Know refer to Distinctive Dual-Domain teachers,  zigzag learning Across RGB-Thermal domains and Incorporating Knowledge from teacher Models.}
  \label{tab:03}
  \vspace{-3mm}
\end{table}

\begin{figure}[t]
  \centering
   \includegraphics[width=1.0\linewidth]{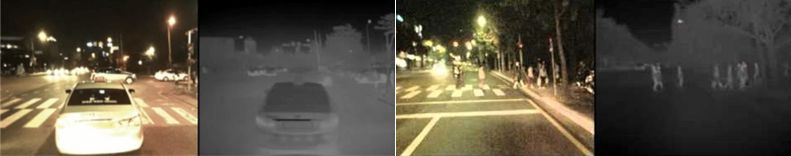}  
   \vspace{-3mm}
   \caption{KAIST RGB and thermal images illustrating disparities between the two domains. Evaluation is conducted without the use of this paired information.}
    \label{fig:rebuttal_01}
   \vspace{-3mm}
\end{figure}

\textbf{Gap between RGB and Thermal}:
We provide both qualitative and quantitative evidences to support our motivation on the large domain gap between RGB and thermal domains. Fig.~\ref{fig:rebuttal_01}, featuring KAIST thermal images, distinctly highlights the unique characteristics of different sensors. Table~\ref{tab:04} shows a substantial performance gap, with results showing 34.68\% for RGB and 65.04\% for thermal oracles, providing quantitative evidence for our motivation. 

\begin{table}[t]
  \centering
  \small
  \begin{tabular}{cc|c}
    \hline
    \multicolumn{2}{c|}{Oracle}& Ours     \\ 
    RGB only & Thermal only & \\ 
     \hline\hline
     34.68 & 65.04 &  69.30  \\
     \hline
    \end{tabular}
  \vspace{-3mm}
  \caption{Comparison of our D3T on FLIR dataset from RGB images to thermal images with the oracle single FCOS models.}
  \label{tab:04}
  \vspace{-3mm}
\end{table}

\textbf{Effect of zigzag learning across RGB-thermal gap:} Table~\ref{tab:05} presents the outcomes of employing zigzag learning across RGB-Thermal domains on the FLIR dataset for adapting from RGB to thermal images, utilizing dynamic iteration settings. The `Fix' setting refers to a consistent training regime where each domain is trained for an equal number of iterations, e.g., 100. In contrast, the `zigzag' setting, as detailed in Section~\ref{sec:3.3}, begins with a focus on the RGB domain before progressively shifting emphasis towards the thermal domain. Note that the frequency of teacher selection is dynamically changed as learning advances. The results indicate that the `zigzag' approach yields a superior mAP by 1.02\%, demonstrating its effectiveness over the `Fix' setting method.

\begin{table}[t]
  \centering
  \small
  \begin{tabular}{c|cccc}
    \hline
    Method & Fix  & Fix & Fix & Zigzag\\
    Iteration & 50 & 100 & 1,000 & Dynamic\\
    \hline
    \hline
    mAP & 65.57 & 68.28 & 65.36 & \textbf{69.30}\\
    \hline
  \end{tabular}
  \vspace{-3mm}
  \caption{Comparison of zigzag learning across RGB-Thermal domains on FLIR dataset from RGB images to thermal images with different iteration settings.}
  \label{tab:05}
  \vspace{-3mm}
\end{table}

\textbf{Effect of incorporating knowledge from teacher models:} 
Table~\ref{tab:06} illustrates the impact of employing pseudo labels to enhance learning capabilities and bridge the domain gap between RGB and thermal domains on the FLIR dataset. The table compares the performance of models trained without pseudo labels, with a fixed \(\lambda\) hyperparameter and with a dynamically \(\lambda\) hyperparameter for equation~(\ref{eq:6}). The results reveal that not using pseudo labels results in an mAP of 68.46\%, whereas using pseudo labels that closely resemble real labels with a \(\lambda\) of 1 leads to a notable decrease in performance, dropping it to 55.12\%. A fixed \(\lambda\) of 0.1 improves the mAP to 68.57\%, and a dynamically changing \(\lambda\) of 0$\rightarrow$1 achieves the best mAP at 69.30\%. This suggests that dynamically adjusting the level of pseudo labels usage during training is an effective strategy to alleviate RGB-Thermal gap, as teachers adapt to the target domain in the later stages of training, instilling trust in their accuracy.

\begin{table}[t]
  \centering
  \small
  \begin{tabular}{c|cccc}
    \hline
    Method & Fixed & Fixed & Fixed & Dynamic\\
    $\lambda$ & 0 & 1 & 0.1 & 0$\rightarrow$1\\
    \hline
    \hline
    mAP & 68.46 & 55.12 & 68.57 & \textbf{69.30}\\
    \hline
  \end{tabular}
  \vspace{-3mm}
  \caption{Comparison of incorporating knowledge from teacher models on FLIR dataset from RGB images to thermal images with different $\lambda$ values.}
  \label{tab:06}
  \vspace{-3mm}
\end{table}

\section{Conclusion}
\label{sec:conclusion}
Our research has effectively navigated the challenges of domain adaptation for object detection from RGB to the thermal domain, a task typically constrained by a lack of extensive thermal datasets and significant domain disparities stemming from RGB data. We have put forth the D3T framework, a novel approach that leverages a dual-teacher model coupled with a zigzag learning regimen, meticulously tailored for adapting from RGB to thermal image. This method markedly enhances model performance, enabling smooth transitions and a focused application of domain-specific knowledge. The results highlight the efficacy of our approach. Our method establishes a solid foundation for subsequent innovations in UDA and sets a groundbreaking benchmark for thermal object detection, bolstering applications dependent on trustworthy vision systems across diverse conditions.

\noindent \textbf{Acknowledgement}: This work was partially supported by Hyundai Motor Company and Kia, and supported by Korea IITP grants (IITP-2024-No.RS-2023-00255968, AI Convergence Innovation Human Resources Dev.; RS-2023-00236245, Dev. of Perception/Planning AI SW for Seamless Auton. Driving in Adverse Weather/Unstructured Env.; No.2021-0-02068, AI Innovation Hub) and by NRF grant (NRF-2022R1A2C1091402). W. Hwang is the corresponding author.

{
    \small
    \bibliographystyle{ieeenat_fullname}
    \bibliography{main}
}
\end{document}


\maketitlesupplementary
\section{Class Activation Mapping}
\label{sec:subcam}
Class Activation Map (CAM)~\cite{zhou2016learning} is an algorithm that generates heatmaps highlighting the critical regions in an image for a particular class. Grad-CAM~\cite{selvaraju2017grad} is an improved version of CAM that can be applied across various network architectures. To illustrate the distinctions between the two teacher models in the initial stages, we employ Eigen-CAM~\cite{muhammad2020eigen}, an advanced version of Grad-CAM.

\begin{figure}
  \centering
  \begin{subfigure}[h]{0.236\textwidth}
      \centering
      \includegraphics[width=\textwidth]{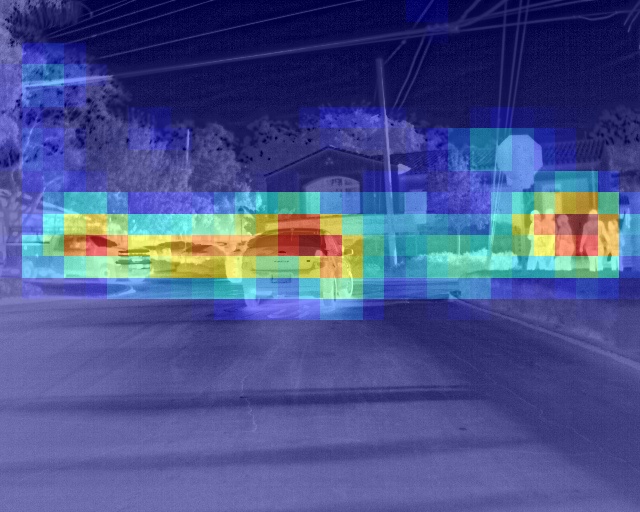}
     \caption{}
  \end{subfigure}
  \hfill
  \begin{subfigure}[h]{0.236\textwidth}
      \centering
      \includegraphics[width=\textwidth]{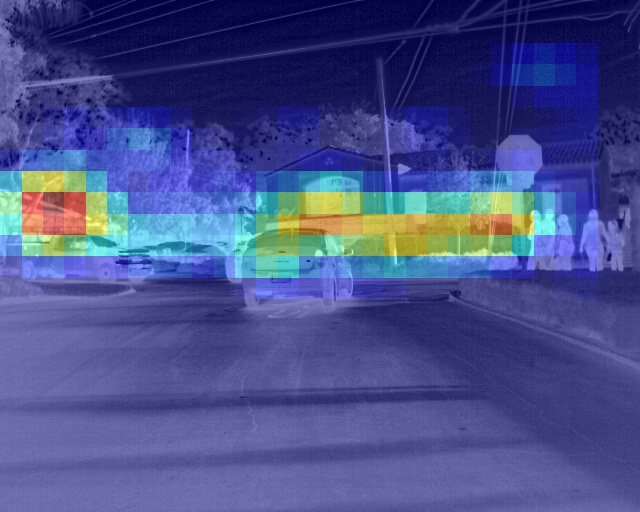}
      \caption{}
  \end{subfigure}
  \hfill
  \begin{subfigure}[h]{0.236\textwidth}
      \centering
      \includegraphics[width=\textwidth]{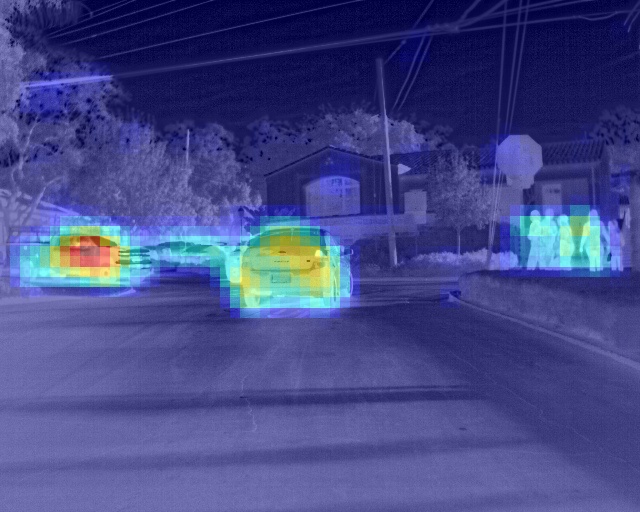}
      \caption{}
  \end{subfigure}
  \hfill
  \begin{subfigure}[h]{0.236\textwidth}
      \centering
      \includegraphics[width=\textwidth]{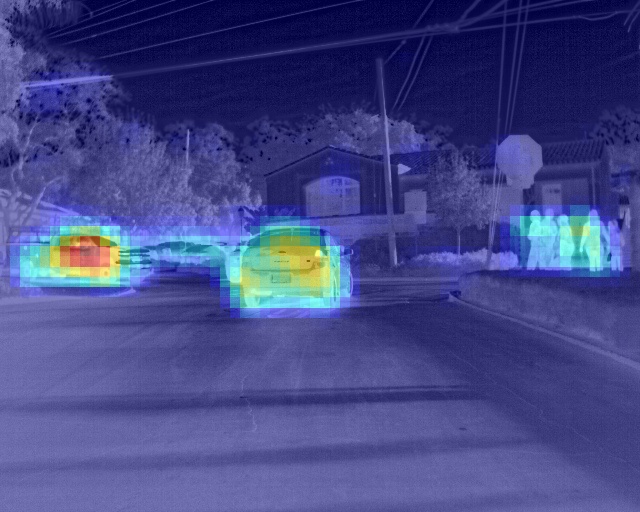}
      \caption{}
  \end{subfigure}
     \caption{CAM generated by dual teachers at various stages of training: (a) and (b) represent CAM from the RGB and thermal teacher models, respectively, during the early stages of training, whereas (c) and (d) are CAM from the same models at later stages of training.}
     \label{fig:sub01}
\end{figure}

As shown in Fig.~\ref{fig:sub01} (a) and (b), during the early stages of training, the RGB teacher and the Thermal teacher show different activation maps. In the later training steps, as shown in Fig.~\ref{fig:sub01} (c) and (d), the activation mappings become similar and focus more on the object. This demonstrates the effectiveness of our proposed method in narrowing the gap between the RGB and Thermal domains, as well as enhancing the accuracy of object recognition.

\section{Performance Changes Across Iterations}
\begin{figure}[t]
  \centering
   \includegraphics[width=1.0\linewidth]{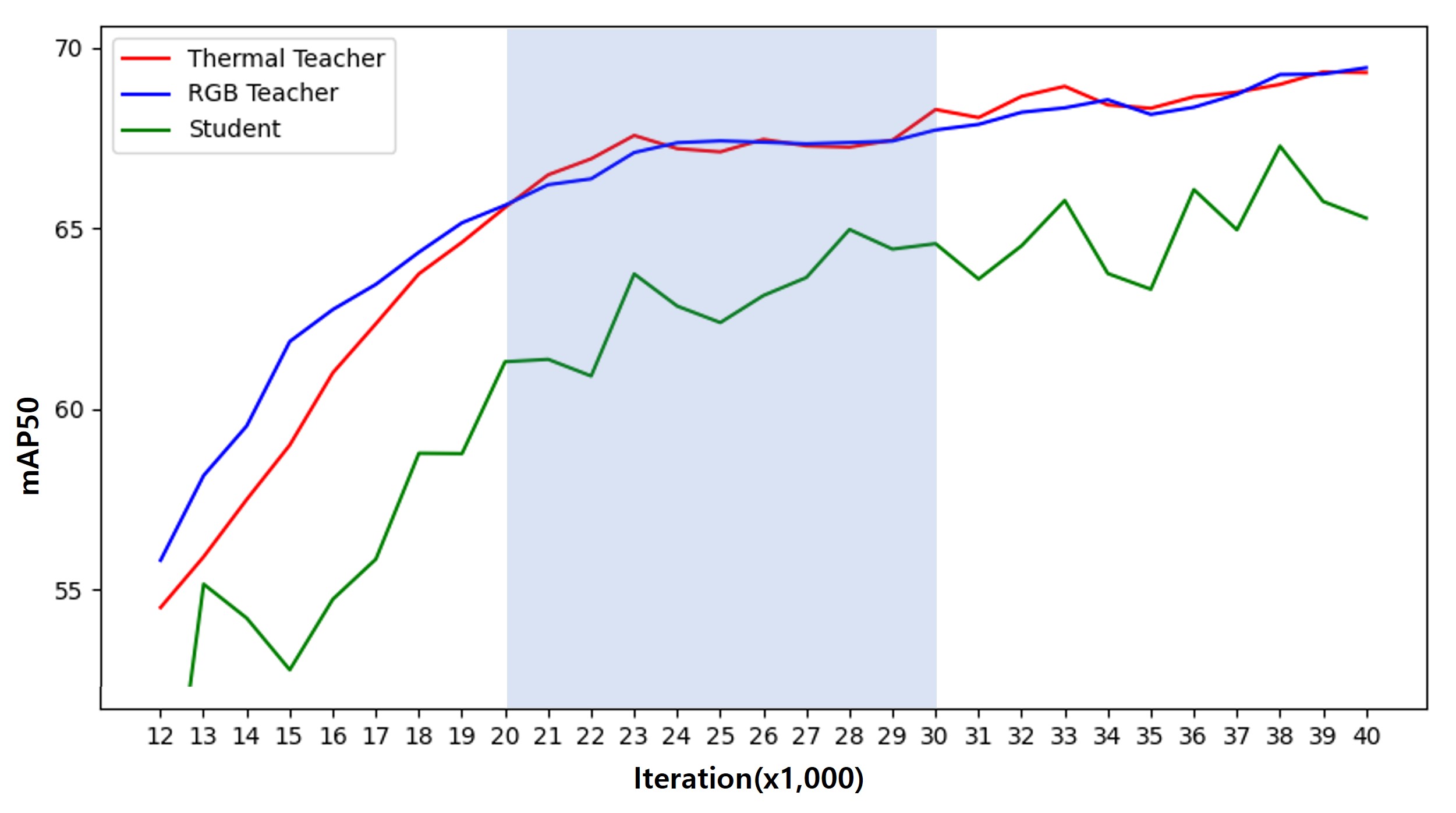}  
   \vspace{-3mm}
   \caption{Performance changes.}
    \label{fig:subloss}
   \vspace{-3mm}
\end{figure}

We plot performance changes of RGB teacher, thermal teacher, and a student across iterations, representing an optimization trend in Fig.~\ref{fig:subloss}. The zig-zag patterns are changed at 20k and 30k iterations. In early stages, RGB teacher outperforms thermal teacher, because we provide more training chances to RGB teacher with labels, then reverses in the middle stages. In the final, their performances converge to similarity, indicating a reduced domain gap. We also note fluctuations in the student's performance, ultimately showing improvement in the upper-right direction. This result demonstrates the stability of our method and we only use thermal teacher for inference. 

\section{RGB DA Benchmark}
We present a new experiment in Table~\ref{tab:sub03}.Our performance has higher mAP than the latest work (e.g., HT~\cite{deng2023harmonious}, AT~\cite{li2022cross}). \textit{Note that our primary focus is on adapting from RGB domain to thermal domain, rather than within RGB domain itself}. 

\begin{table}[t]
  \centering
  \small
  \begin{tabular}{ccc}
    \hline
     HT & AT & Ours \\ 
     \hline\hline
     50.4& 50.9& 51.7  \\
     \hline
    \end{tabular}
  \vspace{-3mm}
  \caption{Cityscapes~\cite{cordts2016cityscapes}$\rightarrow$Foggy Cityscapes~\cite{sakaridis2018semantic}.}
  \label{tab:sub03}
  \vspace{-3mm}
\end{table}
\section{Effect of Zigzag-Learn}
We conduct the ablation test for checking the performance improvement (+1.38\%) of Zigzag-Learn in Table~\ref{tab:subAblation}.

\begin{table}[t]
  \centering
  \small
  \begin{tabular}{ccc|c}
    \hline
     Dual-Teachers & Zigzag-Learn& Incor-Know & mAP \\ 
     \hline\hline
     \checkmark&   & \checkmark & 67.92  \\
     \checkmark & \checkmark & \checkmark & 69.30\\
     \hline
    \end{tabular}
  \vspace{-3mm}
  \caption{Ablation test for components of the proposed method.}
  \label{tab:subAblation}
  \vspace{-3mm}
\end{table}
\section{Novel Approach in Domain Adaptation with Dual Teachers}
The existing methods~\cite{li2022cross, deng2021unbiased} are all based on a single teacher and a single student network for DA. However, we use two teachers, with a student sequentially adopting to each teacher from RGB and thermal domains. \textit{Note that a simple extension from a single to dual teachers does not guarantee optimal performance.} To solve this concern, we propose our novel zigzag-learn and incor-know components. We also conduct an ablation test, comparing ours with AT~\cite{li2022cross} in RGB$\rightarrow$thermal DA test. Table~\ref{tab:sub02} shows that a single teacher-based method fails to achieve satisfactory performance.

\begin{table}[t]
  \centering
  \small
  \begin{tabular}{cc}
    \hline
     AT~\cite{li2022cross}& Ours     \\ 
     \hline\hline
     61.90&  69.30  \\
     \hline
    \end{tabular}
  \vspace{-3mm}
  \caption{RGB$\rightarrow$Thermal in FLIR.}
  \label{tab:sub02}
  \vspace{-3mm}
\end{table}

\section{Pseudo-label Selection}
\begin{figure}[t]
  \centering
   \includegraphics[width=1.0\linewidth]{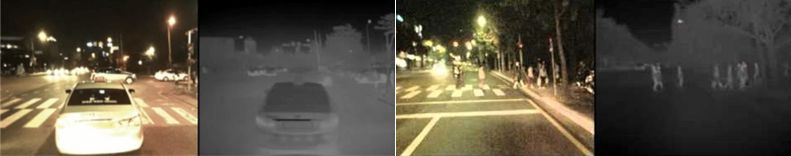}  
   \vspace{-3mm}
   \caption{KAIST RGB and thermal images illustrating disparities between the two domains. Evaluation is conducted without the use of this paired information.}
    \label{fig:subrebuttal_01}
   \vspace{-3mm}
\end{figure}

To solve bad impact of false positives/negatives on pseudo-label, we only employ pseudo-labels in later iterations, once the model has attained stability with dynamically changing $\lambda$ values. We then select only top 1\% pseudo-labels based on confidence values for training the model. As shown in Fig.~\ref{fig:subrebuttal_01}, each modality exhibits unique characteristics, and there is a possibility that labels given from RGB may not align perfectly with thermal labels. This is the primary rationale for employing pseudo-labels to improve the performance in this paper.
\section{Changing $\lambda$}
From iteration 10k to 20k, $\lambda$ gradually increases from 0 to 1 according to the equation (Iteration - 10k)/10k. $\lambda$ equals 1 for iterations greater than 20k and 0 for iterations less than 10k. 
\section{Visualization}
To offer a clearer understanding of the D3T algorithm's effectiveness, we present further results using images from both the source-only model and the D3T model. We present results using both the FLIR and KAIST datasets. The results indicate that our algorithm significantly outperforms the source-only model, which does not utilize the UDA.

\begin{figure}
  \centering
  \begin{subfigure}[h]{0.23633\textwidth}
      \centering
      \includegraphics[width=\textwidth]{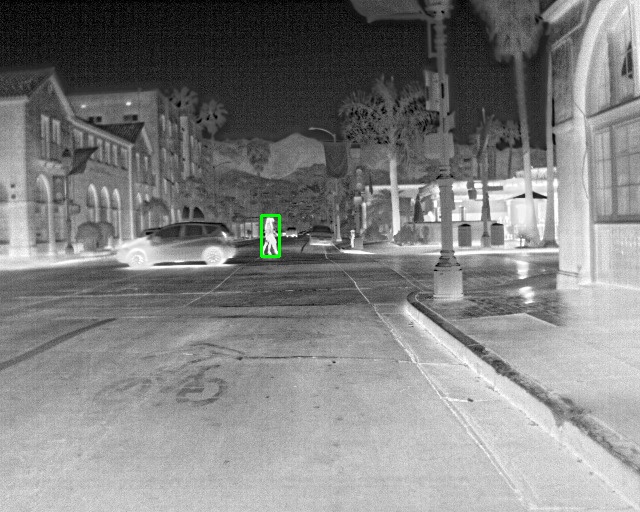}
  \end{subfigure}
  \hfill
  \begin{subfigure}[h]{0.23633\textwidth}
      \centering
      \includegraphics[width=\textwidth]{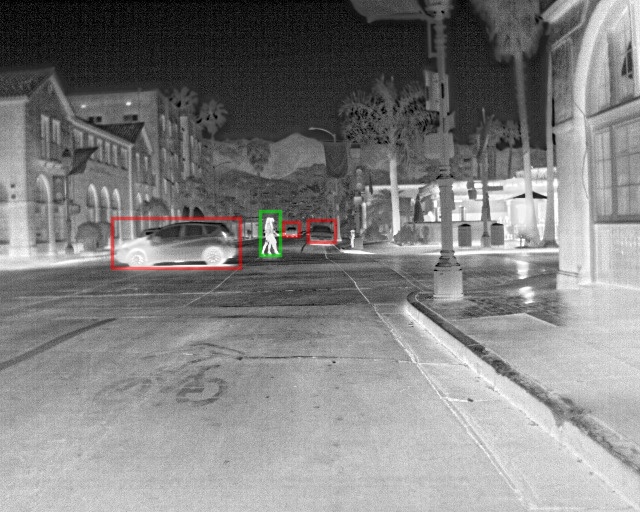}
  \end{subfigure}
  \hfill
  \begin{subfigure}[h]{0.23633\textwidth}
      \centering
      \includegraphics[width=\textwidth]{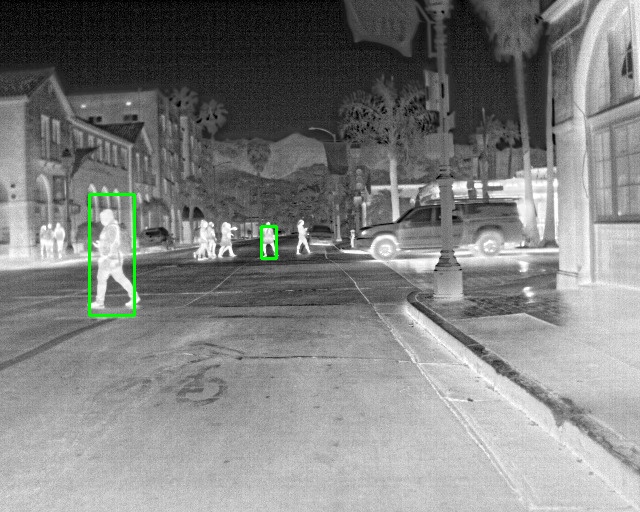}
  \end{subfigure}
  \hfill
  \begin{subfigure}[h]{0.23633\textwidth}
      \centering
      \includegraphics[width=\textwidth]{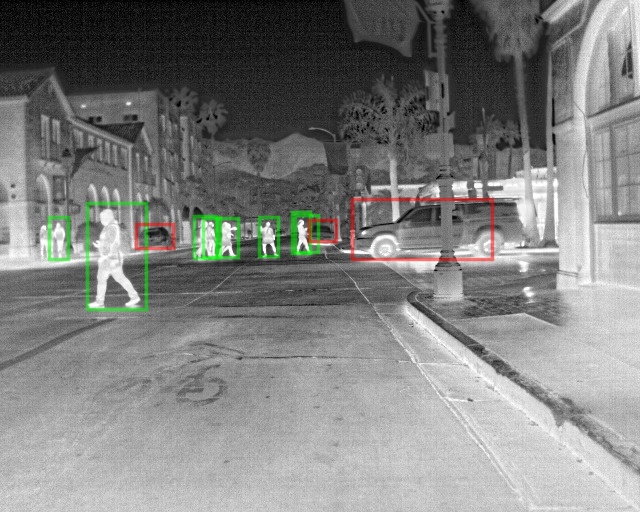}
  \end{subfigure}
  \hfill
  \begin{subfigure}[h]{0.23633\textwidth}
      \centering
      \includegraphics[width=\textwidth]{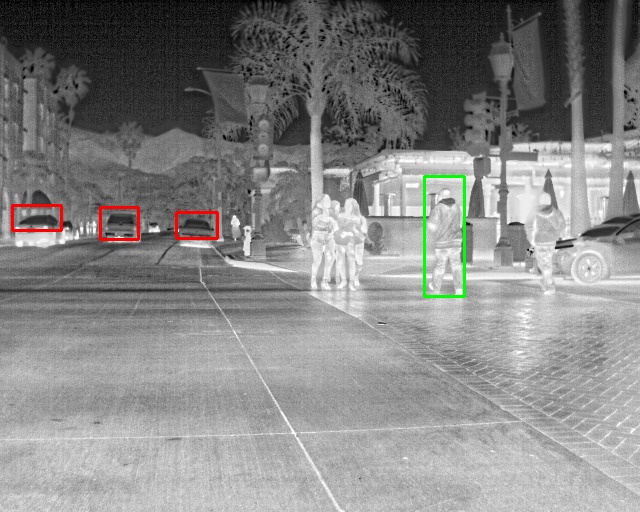}
  \end{subfigure}
  \hfill
  \begin{subfigure}[h]{0.23633\textwidth}
      \centering
      \includegraphics[width=\textwidth]{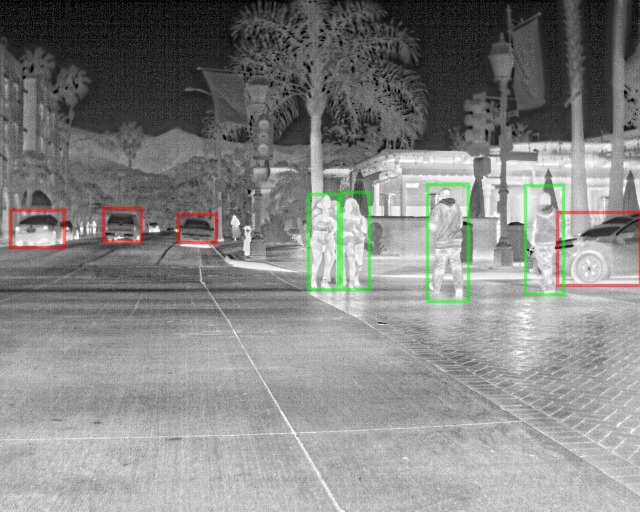}
  \end{subfigure}
  \hfill
  \begin{subfigure}[h]{0.23633\textwidth}
      \centering
      \includegraphics[width=\textwidth]{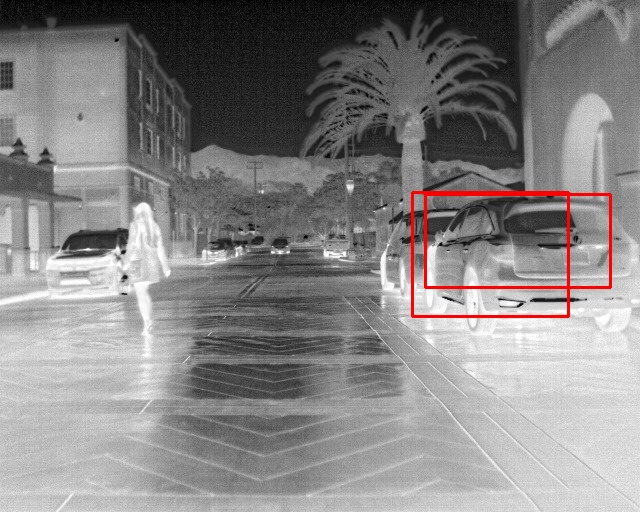}
  \end{subfigure}
  \hfill
  \begin{subfigure}[h]{0.23633\textwidth}
      \centering
      \includegraphics[width=\textwidth]{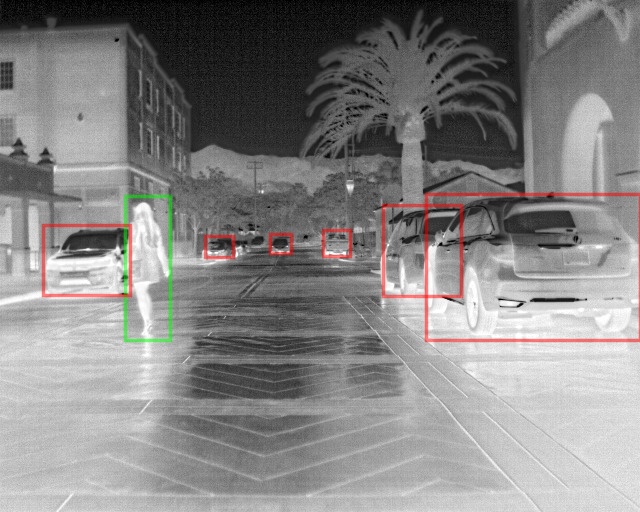}
  \end{subfigure}
  \hfill
  \begin{subfigure}[h]{0.23633\textwidth}
      \centering
      \includegraphics[width=\textwidth]{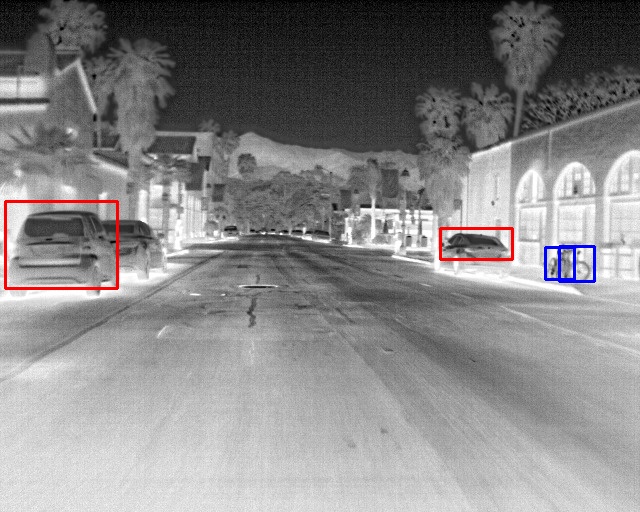}
  \end{subfigure}
  \hfill
  \begin{subfigure}[h]{0.23633\textwidth}
      \centering
      \includegraphics[width=\textwidth]{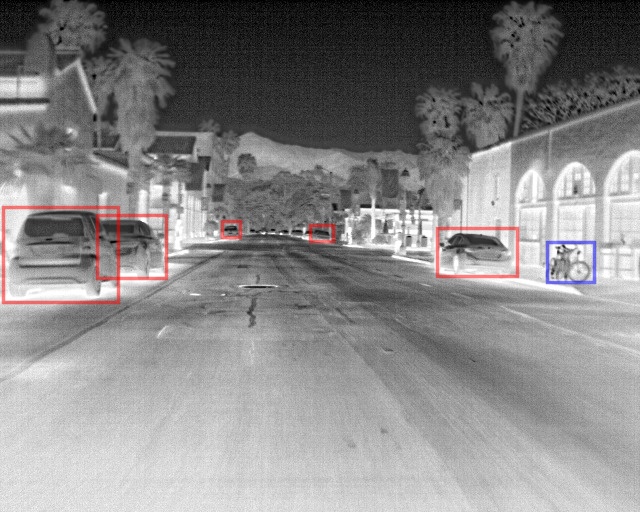}
  \end{subfigure}
  \hfill
  \begin{subfigure}[h]{0.23633\textwidth}
      \centering
      \includegraphics[width=\textwidth]{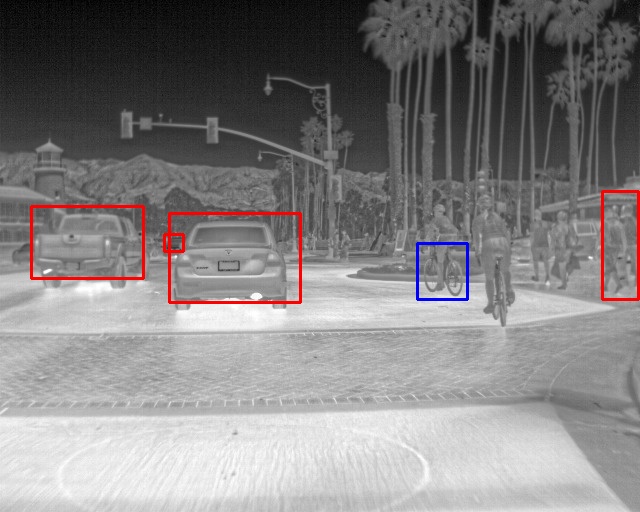}
      \caption{Source only}
  \end{subfigure}
  \hfill
  \begin{subfigure}[h]{0.23633\textwidth}
      \centering
      \includegraphics[width=\textwidth]{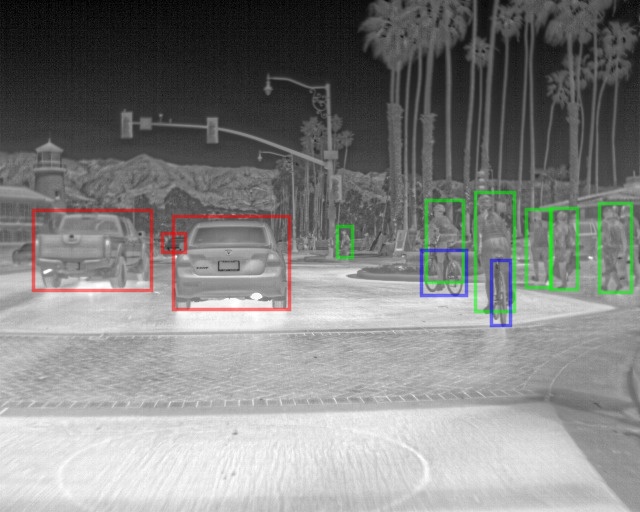}
      \caption{D3T (Ours)}

  \end{subfigure}
     \caption{Visualization of UDA results for object detection models on the FLIR dataset for the RGB to thermal domain: Source-only model and our D3T model. The \textcolor{green}{green}, \textcolor{blue}{blue} and \textcolor{red}{red} boxes represent the classes of \textcolor{green}{person}, \textcolor{blue}{bicycle} and \textcolor{red}{car}.}
     \label{fig:sub02}
\end{figure}

\begin{figure}
  \centering
  \begin{subfigure}[h]{0.23633\textwidth}
      \centering
      \includegraphics[width=\textwidth]{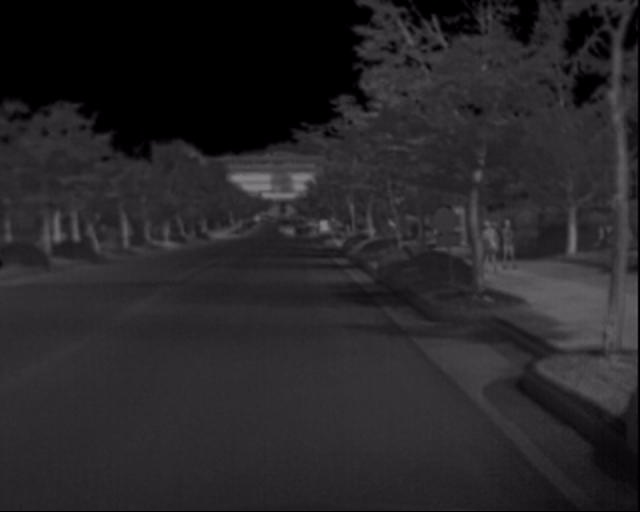}
  \end{subfigure}
  \hfill
  \begin{subfigure}[h]{0.23633\textwidth}
      \centering
      \includegraphics[width=\textwidth]{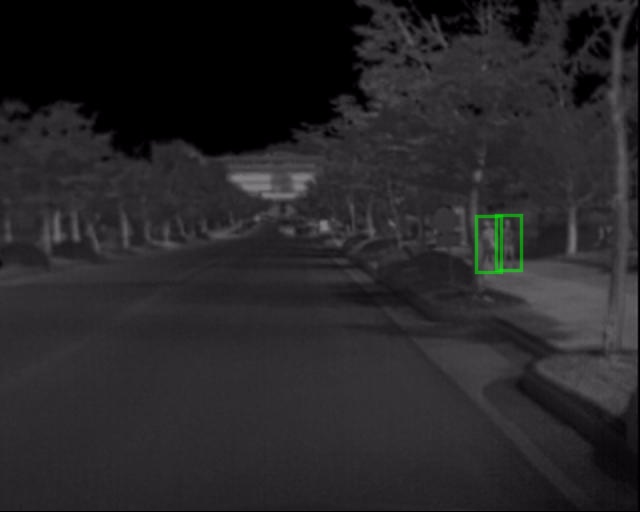}
  \end{subfigure}
  \hfill
  \begin{subfigure}[h]{0.23633\textwidth}
      \centering
      \includegraphics[width=\textwidth]{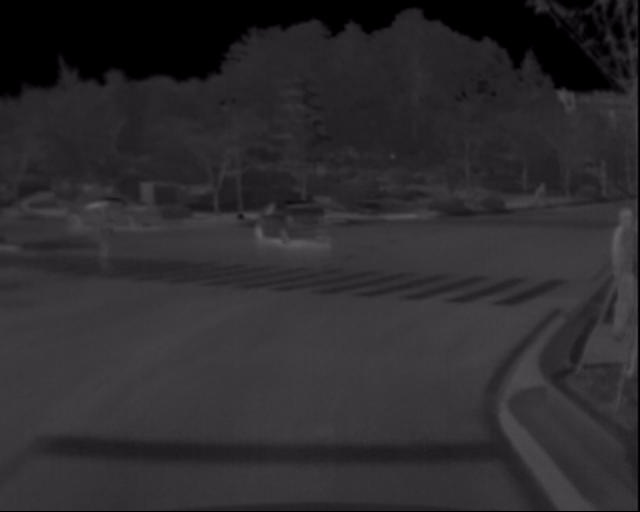}
  \end{subfigure}
  \hfill
  \begin{subfigure}[h]{0.23633\textwidth}
      \centering
      \includegraphics[width=\textwidth]{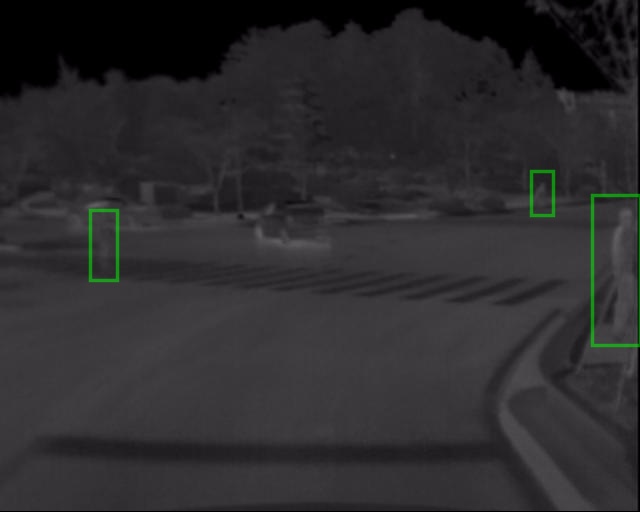}
  \end{subfigure}
  \hfill
  \begin{subfigure}[h]{0.23633\textwidth}
      \centering
      \includegraphics[width=\textwidth]{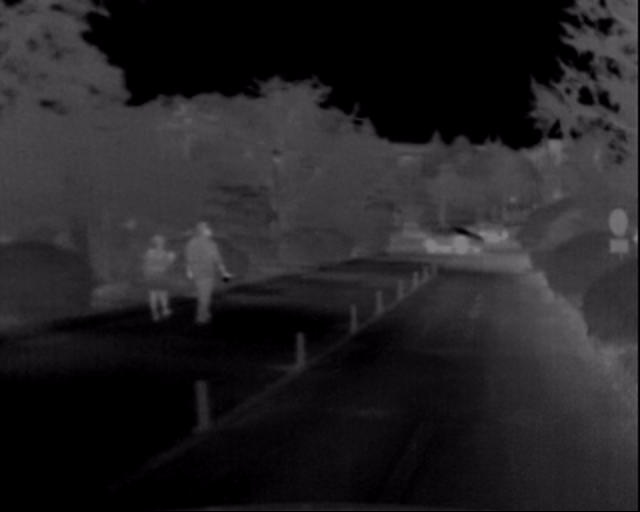}
  \end{subfigure}
  \hfill
  \begin{subfigure}[h]{0.23633\textwidth}
      \centering
      \includegraphics[width=\textwidth]{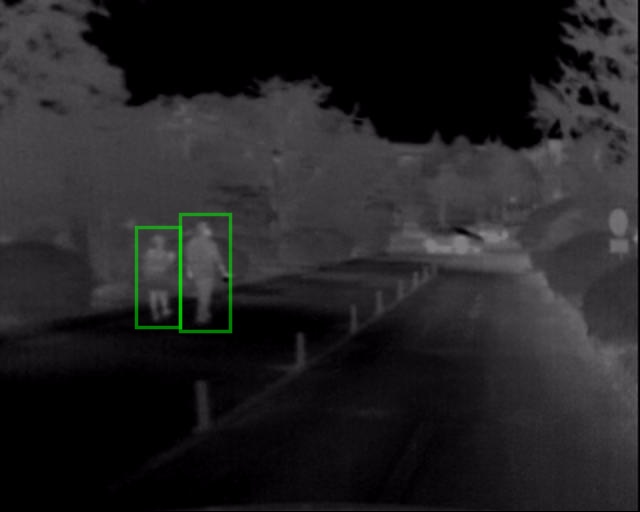}
  \end{subfigure}
  \hfill
  \begin{subfigure}[h]{0.23633\textwidth}
      \centering
      \includegraphics[width=\textwidth]{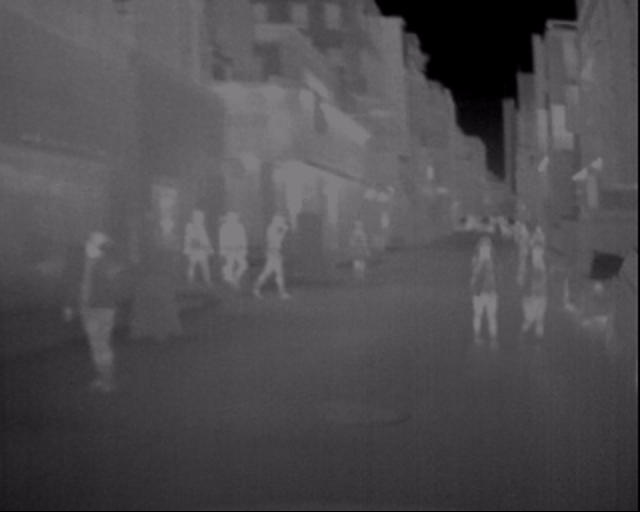}
  \end{subfigure}
  \hfill
  \begin{subfigure}[h]{0.23633\textwidth}
      \centering
      \includegraphics[width=\textwidth]{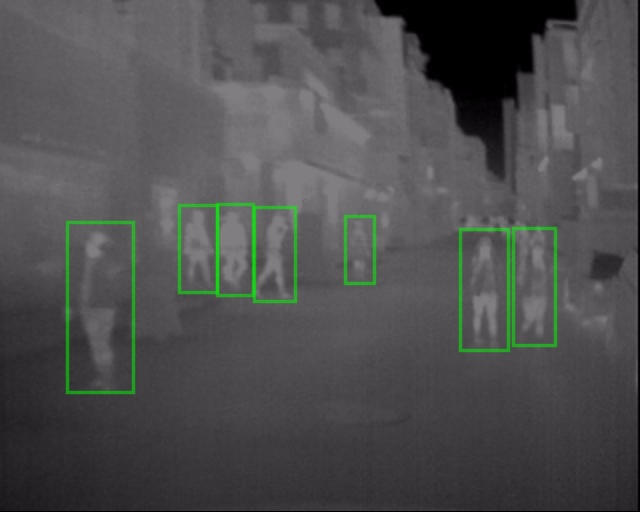}
  \end{subfigure}
  \hfill
  \begin{subfigure}[h]{0.23633\textwidth}
      \centering
      \includegraphics[width=\textwidth]{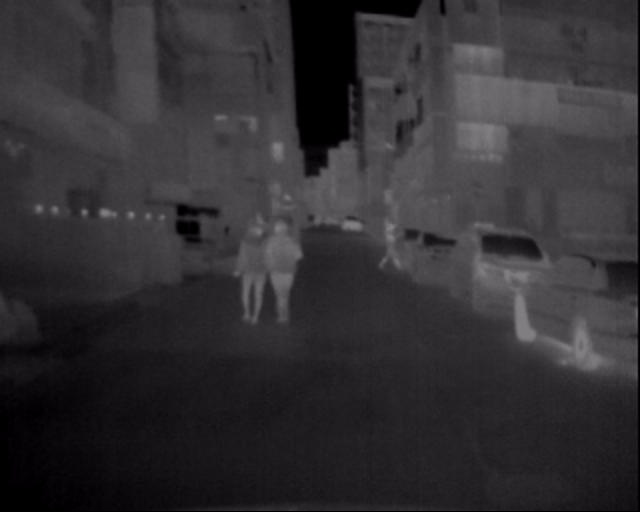}
  \end{subfigure}
  \hfill
  \begin{subfigure}[h]{0.23633\textwidth}
      \centering
      \includegraphics[width=\textwidth]{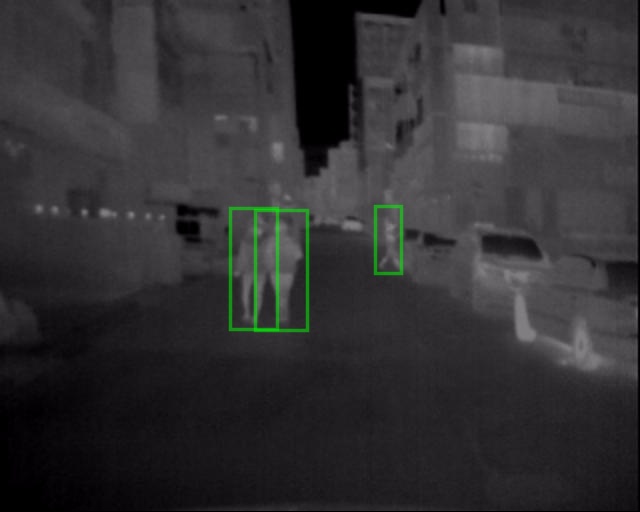}
  \end{subfigure}
  \hfill
  \begin{subfigure}[h]{0.23633\textwidth}
      \centering
      \includegraphics[width=\textwidth]{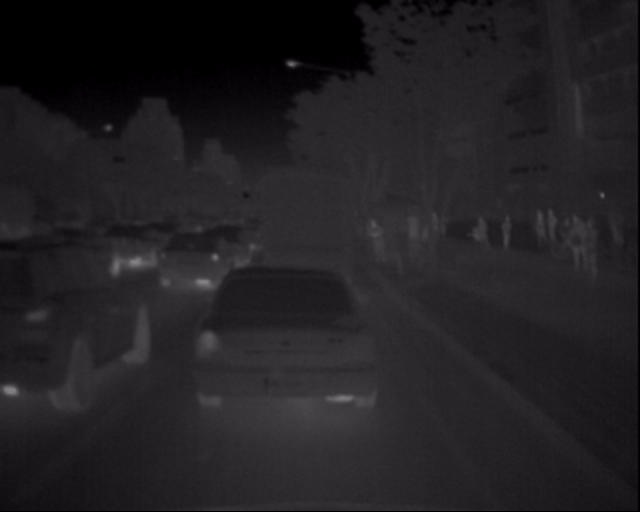}
      \caption{Source only}
  \end{subfigure}
  \hfill
  \begin{subfigure}[h]{0.23633\textwidth}
      \centering
      \includegraphics[width=\textwidth]{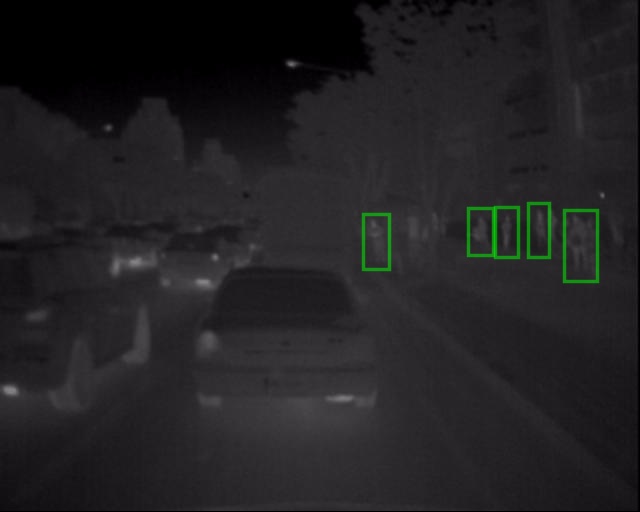}
      \caption{D3T (Ours)}

  \end{subfigure}
     \caption{Visualization of UDA results for object detection models on the KAIST dataset for the RGB to thermal domain: Source-only model and our D3T model. The \textcolor{green}{green} boxes represent the classes of \textcolor{green}{person}.}
     \label{fig:sub03}
\end{figure}
{
    \small
    \bibliographystyle{ieeenat_fullname}
    \bibliography{main}
}